\documentclass[letterpaper, 10 pt, conference]{ieeeconf}  

\IEEEoverridecommandlockouts                              

\usepackage{multirow}
\usepackage{graphicx}
\usepackage{physics}
\usepackage{amsmath}
\usepackage{cite}
\usepackage{amsmath}
\usepackage{algorithm}
\usepackage{algorithmic}
\usepackage{graphicx}
\usepackage{multirow}
\usepackage{booktabs}
\usepackage{amssymb}
\usepackage[hidelinks]{hyperref}
\usepackage{cleveref}
\usepackage{changepage}
\usepackage{subcaption}
\DeclareGraphicsExtensions{.pdf,.jpeg,.png}
\usepackage{pifont}
\usepackage{threeparttable}
\usepackage{textcomp}

\usepackage{amssymb}  

\title{\LARGE \bf Vibration-Assisted Hysteresis Mitigation for Achieving High Compensation Efficiency}

\author{Myeongbo Park$^*$, Chunggil An$^*$, Junhyun Park$^*$, Jonghyun Kang, and Minho Hwang$^{\dagger}$
\thanks{$^*$ These authors are equally contributed.}
\thanks{${\dagger}$ Corresponding Author}
\thanks{Myeongbo Park, Chunggil An, Junhyun Park, Jonghyun Kang, and Minho Hwang are with the Department of Robotics and Mechatronics Engineering, DGIST, Daegu, 42988 Korea (e-mail: {\tt\footnotesize \{qkraudqh23, cndrlfwlq, sean05071, yuri7579, minho\} @dgist.ac.kr}).}
}
\begin{document}

\maketitle
\thispagestyle{empty}
\pagestyle{empty}

\begin{abstract}
Tendon-sheath mechanisms (TSMs) are widely used in minimally invasive surgical (MIS) applications, but their inherent hysteresis—caused by friction, backlash, and tendon elongation—leads to significant tracking errors. Conventional modeling and compensation methods struggle with these nonlinearities and require extensive parameter tuning. To address this, we propose a vibration-assisted hysteresis compensation approach, where controlled vibrational motion is applied along the tendon’s movement direction to mitigate friction and reduce dead zones. Experimental results demonstrate that the exerted vibration consistently reduces hysteresis across all tested frequencies, decreasing RMSE by up to 23.41\% (from 2.2345 mm to 1.7113 mm) and improving correlation, leading to more accurate trajectory tracking. When combined with a Temporal Convolutional Network (TCN)-based compensation model, vibration further enhances performance, achieving an 85.2\% reduction in MAE (from 1.334 mm to 0.1969 mm). Without vibration, the TCN-based approach still reduces MAE by 72.3\% (from 1.334 mm to 0.370 mm) under the same parameter settings. These findings confirm that vibration effectively mitigates hysteresis, improving trajectory accuracy and enabling more efficient compensation models with fewer trainable parameters. This approach provides a scalable and practical solution for TSM-based robotic applications, particularly in MIS.
\end{abstract}

\begin{figure}[t!]
    \centering   
    \vspace{2mm}
    \includegraphics[width=\linewidth]{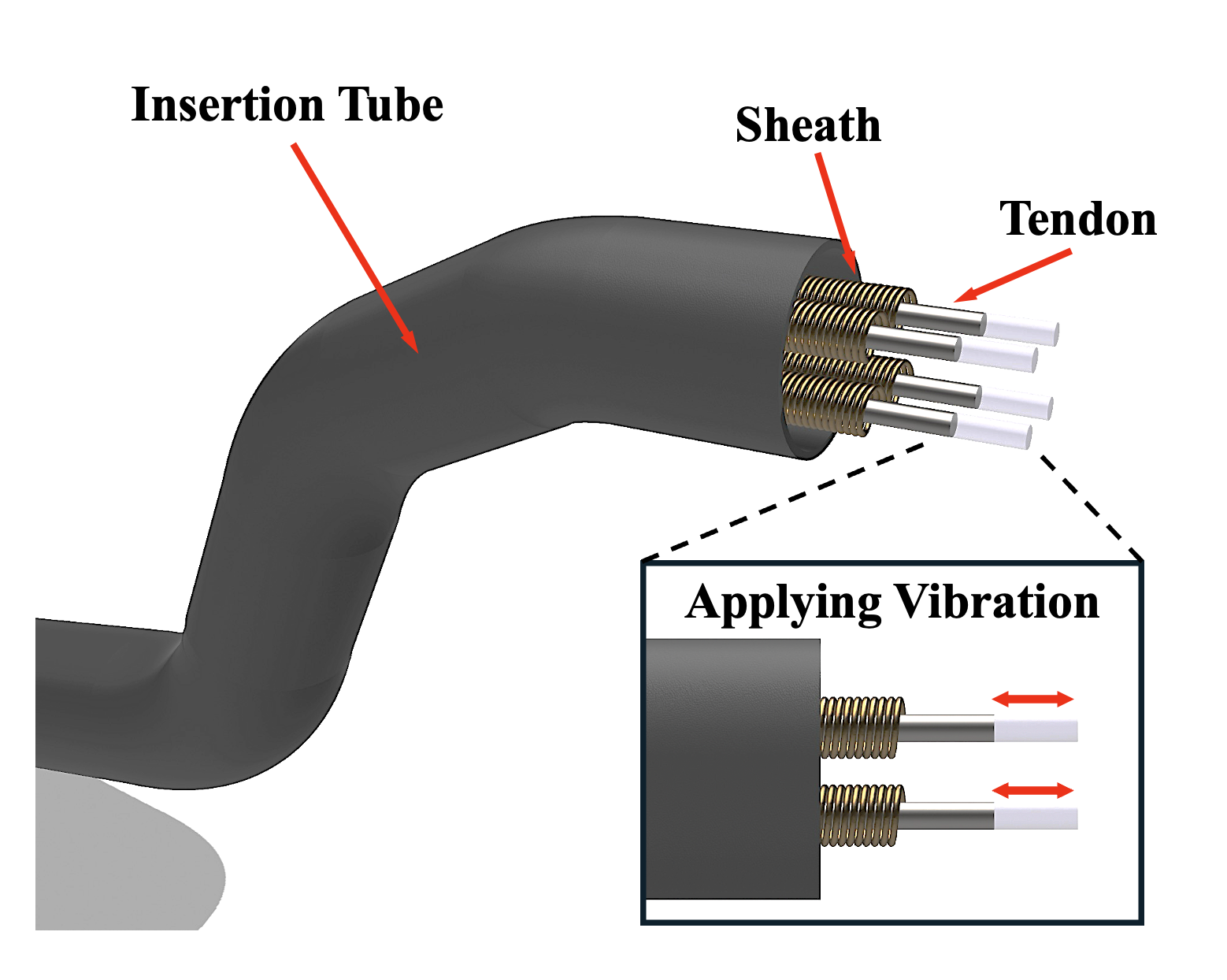}
    \caption{Illustration of the tendon-sheath mechanism (TSM) with applied vibration. The enlarged view highlights the controlled vibrational motion along the tendons, aimed at reducing friction and hysteresis. }
    \label{fig:example}
\end{figure}

\section{INTRODUCTION}

During recent decades, robotics has played an increasingly vital role in medical applications, particularly minimally invasive surgery (MIS)\cite{dvrk}\cite{Strong_K-FLEX}. Unlike rigid surgical robots that navigate lesions through laparoscopic approaches, flexible endoscopic surgical robots utilize tendon-driven mechanisms (TDM) to access endoluminal regions by traversing curved paths through natural orifices such as the mouth, anus, and vagina. Due to their enhanced accessibility \cite{TSMusing1} and the clinical advantages of scar-free procedures, significant research efforts have been dedicated to developing advanced robotic platforms \cite{Hwang2020KFLEXAF, Remacle2015TransoralRS, Phee2009MasterAS, Zorn2018ANT, Kato, FIORA}.

Despite these advantages, tendon-driven systems suffer from a major drawback—hysteresis. The inherent nonlinear characteristics of tendons result in position deviations, significantly affecting the system's precision. To address this, extensive research has been conducted on hysteresis compensation in tendon-driven robots, employing both analytical methods (e.g., modeling dead zones and backlash) and data-driven strategies \cite{park2024hysteresis, park2024sam, Do2014HysteresisMA, Do2015NonlinearFM, Kato2016TendondrivenCR, Kim2021ShapeadaptiveHC, Hwang2020EfficientlyCC}. However, the presence of high nonlinearities from various factors such as friction, tendon twisting \cite{Ji2020AnalysisOT}, extension \cite{Dalvand2018AnAL}, backlash \cite{Kim2020EffectOB}, and coupling effects \cite{Zeng2021MotionCA} complicates the compensation of tendon-sheath mechanism (TSM) hysteresis \cite{Do2014HysteresisMA}.

Given these complexities, it is essential to model and analyze hysteresis in TSMs by considering its fundamental elements individually. TSMs exhibit intricate frictional interactions and coupling effects between the tendon and the sheath, leading to the development of specialized compensation techniques \cite{Do2015NonlinearFM, Lugre1, Lugre2, Bouc1}. These methods compensate for friction-induced phenomena such as dead zones, stick-slip behavior, and the Stribeck effect. However, existing studies do not provide a fundamental solution for reducing friction in TSMs. Additionally, these models require a large number of parameters, leading to significant computational complexity, particularly in unconstrained scenarios with varying tendon motions. Therefore, a more fundamental approach is needed to mitigate hysteresis caused by friction

An eariler study \cite{Eighth} explored a mechanical vibration-based approach to reducing friction in TSMs. By directly applying vibrations to the tendon, the friction coefficient was reduced by up to 50\%. However, this method lacks a comprehensive analysis of hysteresis effects across various tendon movements and does not demonstrate significant compensation improvements through vibration-induced nonlinearity reduction. In this study, we aim to show that vibration reduces positional error by validating the approach on both sinusoidal and random trajectories. Furthermore, we demonstrate that applying vibration decreases nonlinearities, making the system more predictable and easier to compensate for using a neural network (NN) approach. Through ablation studies, we show that even with fewer trainable parameters, the NN model effectively compensates for positional errors, outperforming larger models without vibration.
The main contributions in this paper are as follows:
\begin{itemize}
    \item We validate that inducing vibration reduces position hysteresis in TSMs through controlled vibrational motion, minimizing nonlinearity. We analyze the impact of different vibration frequencies on random trajectories, showing that vibration decreases RMSE by up to 23.4\% (2.2345 mm → 1.7113 mm).
  
    \item We demonstrate that learning-based modeling using a Temporal Convolutional Network (TCN) effectively compensates for hysteresis effects in TSMs. Under the same model settings, the vibration-assisted approach reduces MAE by 85.2\% (1.334 mm → 0.1969 mm), whereas the non-vibration approach achieves a 72.3\% reduction (1.334 mm → 0.370 mm).  

    \item We show that vibrational motion effectively reduces nonlinearities by minimizing dead zones and friction effects, enabling more efficient modeling with fewer trainable parameters. This vibration-assisted approach can be extended to other modeling techniques and has potential for real-world applications.
    
\end{itemize}

\section{System Architecture and Control}

\subsection{Hardware Configuration} 
In \Cref{fig:hardware_configuration_}-(a), the vibration application hardware consists of two modules: an input module that generates vibration and tendon motion, and an output module that applies the initial tension load. Additionally, force sensors and linear encoders are mounted on both modules to measure tension efficiency. The components of each module are described as follows:

\begin{figure}[t!]
    \centering
    \includegraphics[width=1.0\linewidth]{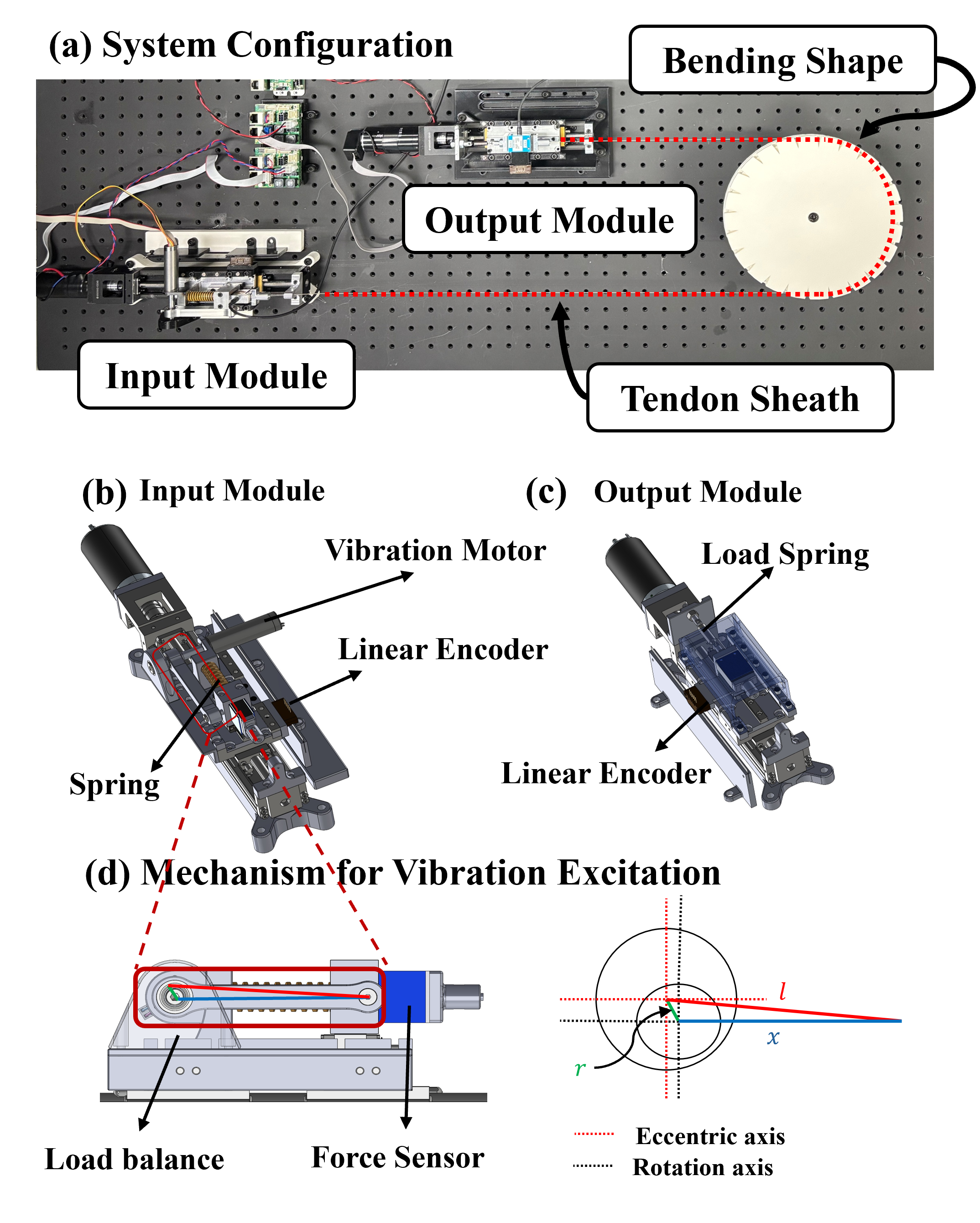}
    \caption{Hardware configuration of the Input and Output Modules.  
     (a) Overall Hardware configuration
     (b) The Input Module includes an eccentric axis, crank slider, vibration motor for tendon vibration.
     (c) The Output Module consists of tension spring for tendon load.
     (d) Crank-slider structure of the vibration mechanism designed to induce vibration in tension.}
    \label{fig:hardware_configuration_}
\end{figure}

\textbf{Input Module}: This module measures the input tension and displacement while generating tendon vibrations and motion as shown in \Cref{fig:hardware_configuration_}-(b). A crank-slider mechanism with an eccentric shaft mounted on the motor shaft induces vibration, as illustrated in \Cref{fig:hardware_configuration_}-(d). The crank length is set to 7.2 mm, and the eccentricity to 0.2 mm, enabling the crank block to perform linear motion, as described by \eqref{eqn:1} and shown in \Cref{fig:hardware_configuration_}-(d). The displacement of the crank-slider mechanism can be formulated as follows:
\begin{equation}
\begin{aligned}
    x &= r \cos(\theta) + \sqrt{l^2 - r^2 \sin^2(\theta)} \\
    x &\approx l + r \cos(\theta) \quad \text{if} \quad l \gg r
\end{aligned}
\label{eqn:1}
\end{equation}

where $r$ is the eccentricity, $l$ is the link length, and $x$ is the slide displacement length. The eccentricity of 0.2 mm indicates that the outer hole of the eccentric shaft is offset by 0.2 mm from the motor's central axis, which is attached to the inner hole of the eccentric shaft. This mechanism converts the rotation of the vibration motor into longitudinal tendon vibrations. 

The ECX SPEED 19 motor is used as the vibration motor. Additionally, a load balancer is integrated to reduce motor load, and a spring is placed between the force sensor fixture and the motor fixture to mitigate axial loads on the vibration motor when input tension is applied. This system can generate vibrations at frequencies of up to 100 Hz.

\textbf{Output Module}: This module measures both the output tension and displacement while applying initial and variable tension loads to the system, as illustrated in \Cref{fig:hardware_configuration_}-(c). A spring, attached to the end of the force sensor in the output module, serves as an environmental variable to measure the applied tension load and maintain the initial tension. The spring has a stiffness constant of 1.08 N/mm and a maximum force of 30.0 N.

The experimental setup consists of a tendon with a length of 700.0 mm and a diameter of 0.36 mm, enclosed in a sheath with an outer diameter of 1.0 mm and a wire thickness of 0.2 mm. Both input and output linear actuation are driven by Maxon 357962 RE35 DC motors, with motion controlled by Maxon EPOS4 50/5 Compact Motion Controllers. Tension measurements are performed using KTOYO FDX10 axial force sensors, with data acquisition managed by the PHIDGET BRIDGE 1047\_2B. Additionally, displacement is recorded using a US Digital EM2 linear encoder at a sampling rate of 55 Hz.

\begin{figure}[t!]
    \centering
    \vspace{2mm}
    \includegraphics[width=\linewidth]{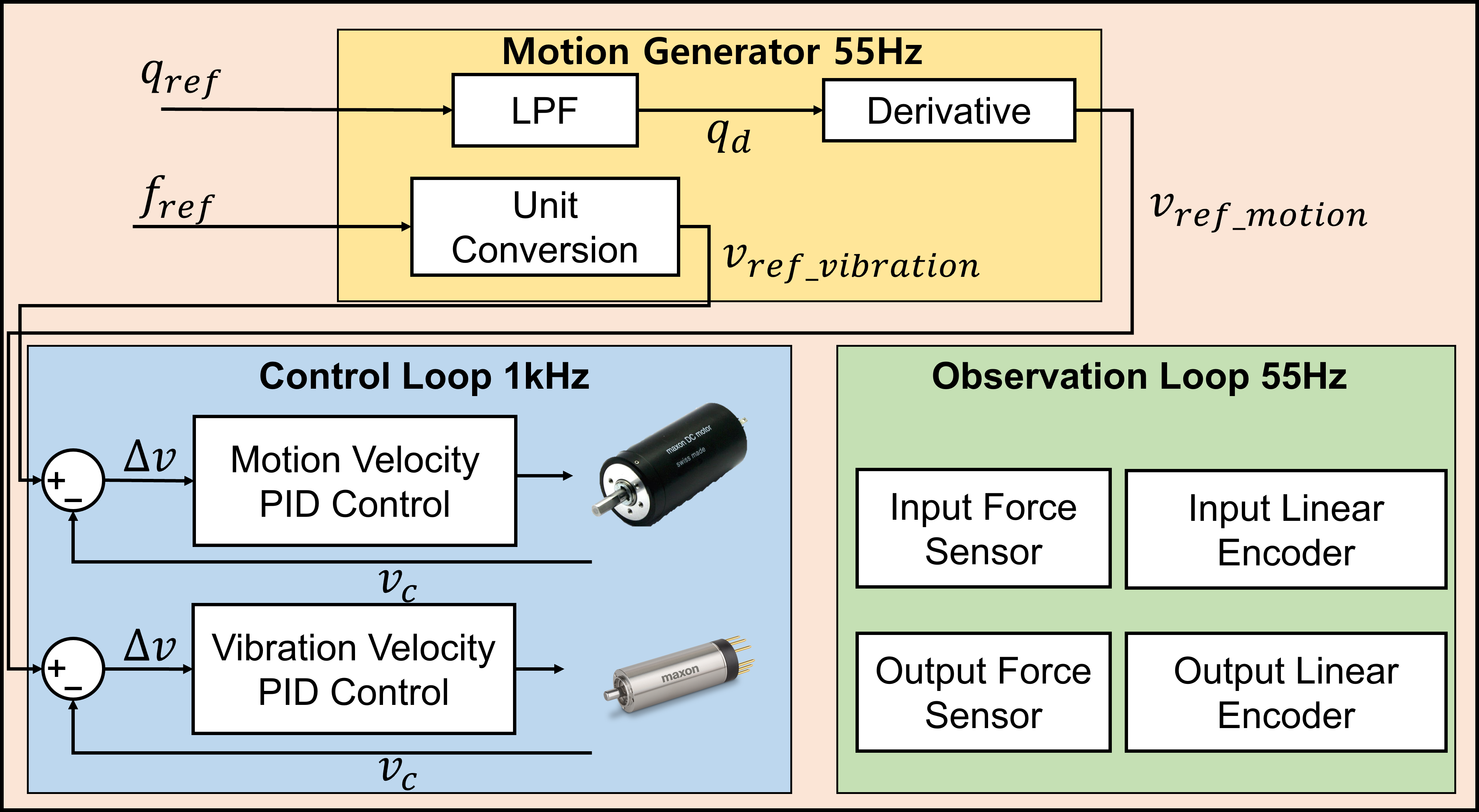} 
    \caption{Control architecture of the TSM system, consisting of a motion generator (55Hz) for trajectory processing, a control loop (1kHz) for motion and vibration control, and an observation loop (55Hz) for real-time system monitoring.}
    \label{fig:control} 
\end{figure}

\subsection{Control Methods}
Figure~\ref{fig:control} illustrates the control architecture designed for the TSM system. The architecture consists of three primary components: the Motion Generator (55 Hz), the Control Loop (1 kHz), and the Observation Loop (55 Hz).

The Motion Generator receives the reference trajectory \( q_{ref} \) as input and generates the desired motion trajectory while incorporating both the primary motion (tendon motion) and the vibration component. To ensure smooth motion generation, a windowed approach is applied using the past 20 samples of \( q_{ref} \):
\begin{equation}
    q_d = \text{LPF}(q_{ref})
\end{equation}
where LPF represents a low-pass filtering (LPF) process that eliminates high-frequency noise and smooths the trajectory. Once the smoothed trajectory \( q_d \) is obtained, a differentiation operation is performed to compute the reference motion velocity:
\begin{equation}
    v_{\text{ref, motion}} = \frac{d q_d}{dt}
\end{equation}
Simultaneously, a vibration velocity component is generated using the frequency reference \( f_{ref} \) and converted into an equivalent velocity command.

Both velocity components are then utilized in the Control Loop, which operates at 1 kHz. This loop consists of two PID velocity controllers: one for tracking the motion velocity \( v_{\text{ref, motion}} \) and another for tracking the vibration velocity \( v_{\text{ref, vibration}} \). These controllers minimize velocity errors and provide actuation commands to the respective motors, ensuring that both the primary motion and the vibration-induced movement are accurately controlled.

The Observation Loop, running at 55 Hz, collects sensor data to assess the system’s response to both motion and vibration inputs. It includes force sensors that measure input and output forces, as well as linear encoders that track the displacement of the tendon mechanism. The collected data allows for quantification of hysteresis effects and evaluation of how vibration influences positional accuracy.

\section{Hysteresis Behavior Observation of TSM}
In this section, we examine the hysteresis behavior of TSM under vibrational motion. First, we analyze the effect of vibration on tendon motion using a sinusoidal trajectory. By varying the initial tension and motion frequency, We investigate the influence of vibration on tendon motion characteristics. Additionally, to evaluate the impact of vibration on a random trajectory (which contains multiple frequency components within a single cycle), we generate a random trajectory and apply different vibration frequencies to the tendon.

\begin{figure*}[t!]
    \centering
    \includegraphics[width=\linewidth]{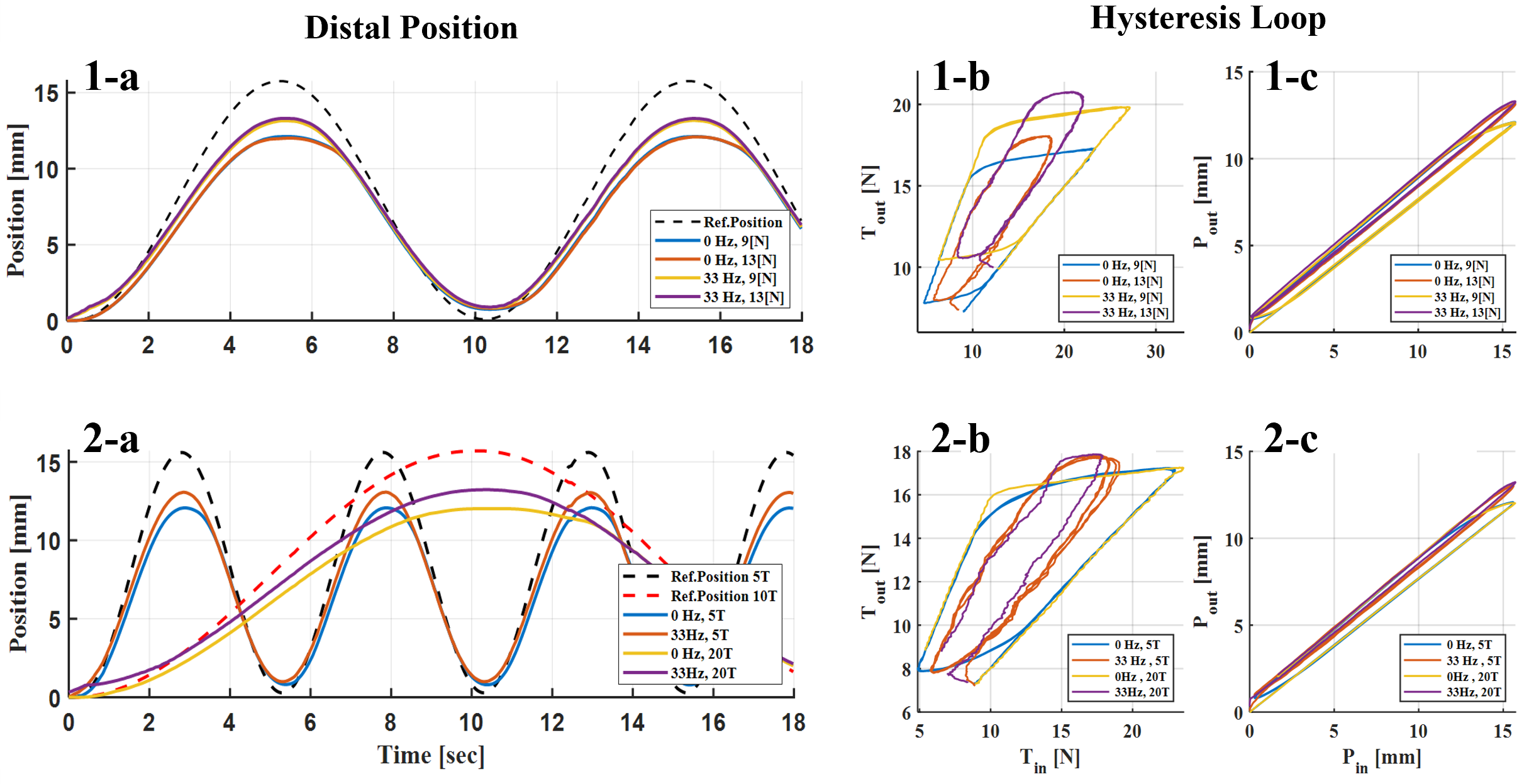}
    \caption{Graph of Hysteresis effect on Sinusoid Trajectory, (1-a) Distal position in the time domain from the initial tension experiment. (1-b) Tension hysteresis observed in the same experiment. (1-c) Hysteresis observed in relation to the position. (2-a) Time-domain graph illustrating the experiment related to frequency and vibration. (2-b) The relationship between input Tension and output Tension. (2-c) Hysteresis loop showing the correlation between input and output positions.}
    \label{fig:sinusoid_analysis}
\end{figure*}

\subsection{TSM Hysteresis Behavior with Vibrational Motion on Sinusoid Trajectory}
This subsection investigates hysteresis behavior by comparing cases with and without vibration, considering two initial tension levels and two tendon motion frequencies. The initial tension and motion frequency were chosen as variables since increasing the initial tension is expected to generate a higher normal force in the $180^\circ$ bending region. Consequently, initial tension is analyzed in relation to friction force and applied vibration frequency. Additionally, motion frequency is introduced as a variable to examine interactions between tendon-driving motion vibrations and externally induced tendon vibrations. The reference tendon motion follows a sinusoidal trajectory with a 16 mm amplitude and period $T$
(set as 1 seconds).

To assess the effect of initial tension in TSM, we varied the initial tension between 9N and 13N while maintaining a constant tendon motion period of $10T$ under vibrational (33 Hz) and non-vibrational (0 Hz) conditions. As shown in \Cref{fig:sinusoid_analysis}.1-a, time-domain analysis reveals that applying vibration reduces the phase lag in the output position due to friction, improving position gain with a maximum reduction rate of 34.6\%. Furthermore, vibration mitigates tension loss at the output stage and reduces the dead zone effect. However, as shown in \Cref{fig:sinusoid_analysis}.1-b and 1-c, initial tension has minimal impact on the system's hysteresis characteristics.

Next, we examine the influence of the tendon motion period under vibrational (33 Hz) and non-vibrational (0 Hz) conditions. The tendon motion period was set to $5T$ and $20T$, while the initial tension remained constant at 8.5 N. As observed in \Cref{fig:sinusoid_analysis}.2-a, vibration increased responsiveness and position gain by 33.79\%, consistent with previous findings. Additionally, as shown in \Cref{fig:sinusoid_analysis}.2-b, the tension hysteresis loop for $20T$ exhibits a narrower distribution compared to $5T$, while the position hysteresis pattern aligns with the initial tension experiment.

These results confirm that vibration effectively reduces dead zone effects and tension loss due to friction in TSMs, leading to reduced position error. While no significant differences were observed in the position hysteresis loop distribution across different initial tensions, the motion frequency experiment revealed a smaller tension hystersis loop distribution for $20T$ in \Cref{fig:sinusoid_analysis}.2-b. This suggests that the inducing vibration on simple harmonic motion can be reduced the position error, and further applicable on random trajectory, which contains multiple frequency components within a single cycle, with more different vibrational frequency induced.

\subsection{Random Trajectory Generation}
\label{sec:random_trajectory}

To investigate the effect of vibration frequency on system behavior in random trajectories, an experiment was conducted by applying various frequencies to an unstructured motion pattern. To enhance randomness and better approximate real-world motion variations, different interpolation methods were applied between randomly sampled waypoints.

The trajectory begins at an initial position of \( p_0 = 0 \) mm, with subsequent waypoints \( p_i \) randomly sampled from a uniform distribution as follows:
\begin{equation}
    p_i \sim \mathcal{U}(5, 15), \quad i = 1, 2, \dots, N-1
\end{equation}

where \( N \) denotes the total number of waypoints. To introduce additional variability, the interpolation method for each segment \( (p_i, p_{i+1}) \) was randomly selected from the following techniques: 1) Linear Interpolation, (Lerp), 2) Cubic Hermite Interpolation, and 3) Catmull-Rom Spline Interpolation.

By combining random waypoint selection with varied interpolation methods, the trajectory was generated in a non-deterministic manner rather than being predefined. This stochastic approach ensures an unbiased analysis of how different vibration frequencies influence the system’s response.

\subsection{Effect of Vibration Frequency on Trajectory Tracking}
To systematically assess the effect of vibration, frequencies from 0 Hz (no vibration) to 100 Hz were applied in increments of 10 Hz to the same randomly generated trajectory. The results, summarized in Table~\ref{tab:result_of_frequency}, evaluate the impact of vibration using RMSE and the Pearson correlation coefficient.  
\begin{equation}
    r = \frac{\sum (p_d - \bar{p}_d) (p_m - \bar{p}_m)}
    {\sqrt{\sum (p_d - \bar{p}_d)^2 \sum (p_m - \bar{p}_m)^2}}
\end{equation}

where \( r \) represents the Pearson correlation coefficient, \( p_d \) and \( p_m \) are the desired and measured trajectories, respectively, and \( \bar{p}_d \) and \( \bar{p}_m \) denote their mean values.
\begin{equation}
    r_{\text{norm}} = \frac{r - \min(r)}{\max(r) - \min(r)}
\end{equation}

From Table~\ref{tab:result_of_frequency}, it is observed that applying vibration at all tested frequencies led to a reduction in RMSE and an increase in correlation coefficient. Compared to the no-vibration condition (0 Hz), RMSE decreased across all tested frequencies, with a maximum reduction of approximately 23.41\% (from 2.2345 mm to 1.7113 mm) observed at 90 Hz. Additionally, the correlation coefficient increased for all vibration frequencies, indicating that the system followed the reference trajectory more closely when vibration was applied, effectively mitigating the dead zone effect.  

\begin{figure}[t!]
    \centering
    \includegraphics[width=1.0\linewidth]{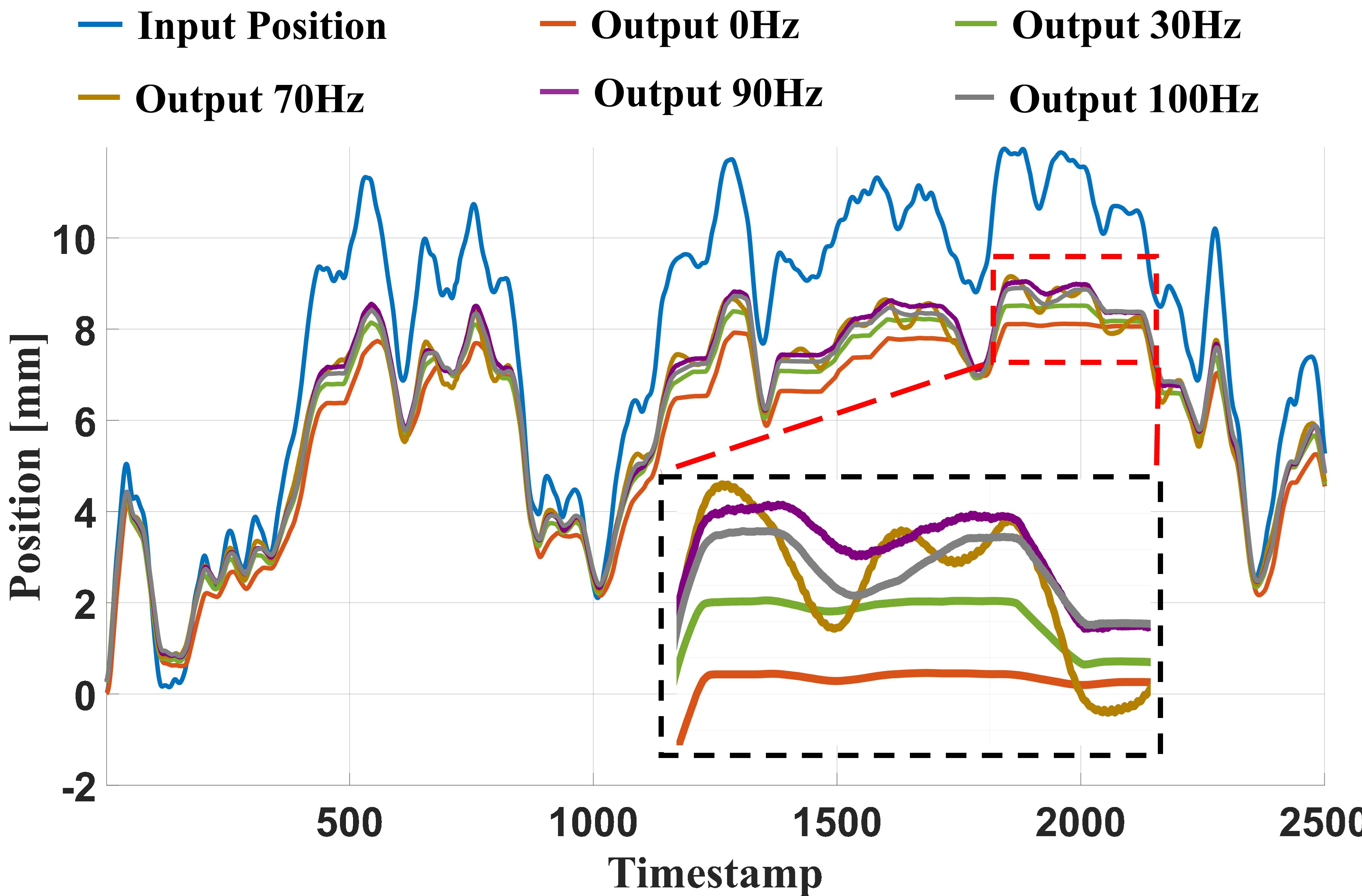} 
    \caption{Trajectory tracking performance under different vibration frequencies. The application of vibration reduces hysteresis and improves tracking accuracy. The black box highlights the reduction in dead zones.}
    \label{fig:random_frequency} 
\end{figure}

Meanwhile, Figure~\ref{fig:random_frequency} provides a qualitative illustration of how vibration helps mitigate dead zones, especially at 70 Hz, where the correlation is the highest. The inset box in Figure~\ref{fig:random_frequency} highlights the reduction in dead zones when vibration is applied, further supporting the role of vibration in improving trajectory tracking accuracy.  

Applying vibration consistently resulted in significant improvements in both RMSE and correlation, with no abrupt variations in performance across different frequencies. While 90 Hz yielded the lowest RMSE, 70 Hz exhibited the highest correlation. Considering that 70 Hz and 90 Hz had comparable RMSE values (difference of 0.08 mm), but 70 Hz showed a greater reduction in dead zones compared to 90 Hz, it resulted in the highest normalized correlation. Therefore, we selected 70 Hz for further TSM modeling experiments.

\begin{table}[t!]
\centering
\caption{This table states the positional hysteresis on each specific frequency applied. }
\label{tab:result_of_frequency}
\resizebox{0.8\linewidth}{!}{%
\begin{tabular}{c c c c} 
\toprule
\textbf{Frequency}         & \begin{tabular}[c]{@{}c@{}}\textbf{RMSE }\\\textbf{[mm]}\end{tabular} & \begin{tabular}[c]{@{}c@{}}\textbf{STD }\\\textbf{[mm]}\end{tabular} & \begin{tabular}[c]{@{}c@{}}\textbf{Normalized}\\\textbf{Correlation}\end{tabular}  \\ 
\midrule
\textbf{0 Hz}  & \textbf{2.2345}  & 1.2090  & \textbf{0.0000}  \\
10 Hz          & 1.9155           & 1.1295  & 0.8061  \\
20 Hz          & 1.9466           & 1.1246  & 0.7665  \\
30 Hz          & 1.9469           & 1.1227  & 0.7828  \\
40 Hz          & 1.9083           & 1.1050  & 0.7277  \\
50 Hz          & 1.9120           & 1.1023  & 0.7548  \\
60 Hz          & 1.9466           & 1.1351  & 0.6206  \\
70 Hz          & 1.7974           & 1.0647  & \textbf{1.0000}  \\
80 Hz          & 1.7561           & 1.0550  & 0.9666  \\
90 Hz          & \textbf{1.7113}  & 1.0309  & 0.7603  \\
100 Hz         & 1.8366           & 1.0958  & 0.7099  \\
\toprule
\end{tabular}
}
\end{table}

\section{Hysteresis Modeling and Compensation}
Hysteresis in TSM is commonly modeled using either model-based or learning-based approaches. In this section, we focus on a learning-based modeling approach and demonstrate that applying vibration to the tendon significantly reduces nonlinearities in hysteresis. This reduction enhances estimation performance while maintaining the same trainable parameters, hyperparameters, and experimental settings.

\subsection{Data Formulation}
To model hysteresis in the TSM, we collected data from a randomized trajectory. The dataset consists of commanded position inputs and their corresponding measured outputs under two conditions: without vibration and with vibration. The datasets are structured as follows:
\begin{align}
\mathcal{D}_{\text{no vib}} = \left \{ \left( \textbf{p}_{\mathrm{cmd}}(t), \textbf{p}_{\mathrm{meas, no \: vib}}(t) \right) \right \}_{t=1}^{N} \\ \nonumber
\mathcal{D}_{\text{vib}} = \left \{ \left( \textbf{p}_{\mathrm{cmd}}(t), \textbf{p}_{\mathrm{meas, vib}}(t) \right) \right \}_{t=1}^{N}
\end{align}

where \(N\) represents the total number of data points, with \(N = 15,000\). To evaluate the effect of vibration on the tendon-sheath mechanism, we applied the same commanded trajectory \(\textbf{p}_{\mathrm{cmd}}(t)\) under two conditions: without vibration, yielding the measured output \(\textbf{p}_{\mathrm{meas, no vib}}(t)\), and with vibration, yielding \(\textbf{p}_{\mathrm{meas, vib}}(t)\). The vibration frequency was maintained at a constant 70 Hz. 

\begin{figure}[t!]
    \centering
    \vspace{2mm}
    \includegraphics[width=\linewidth]{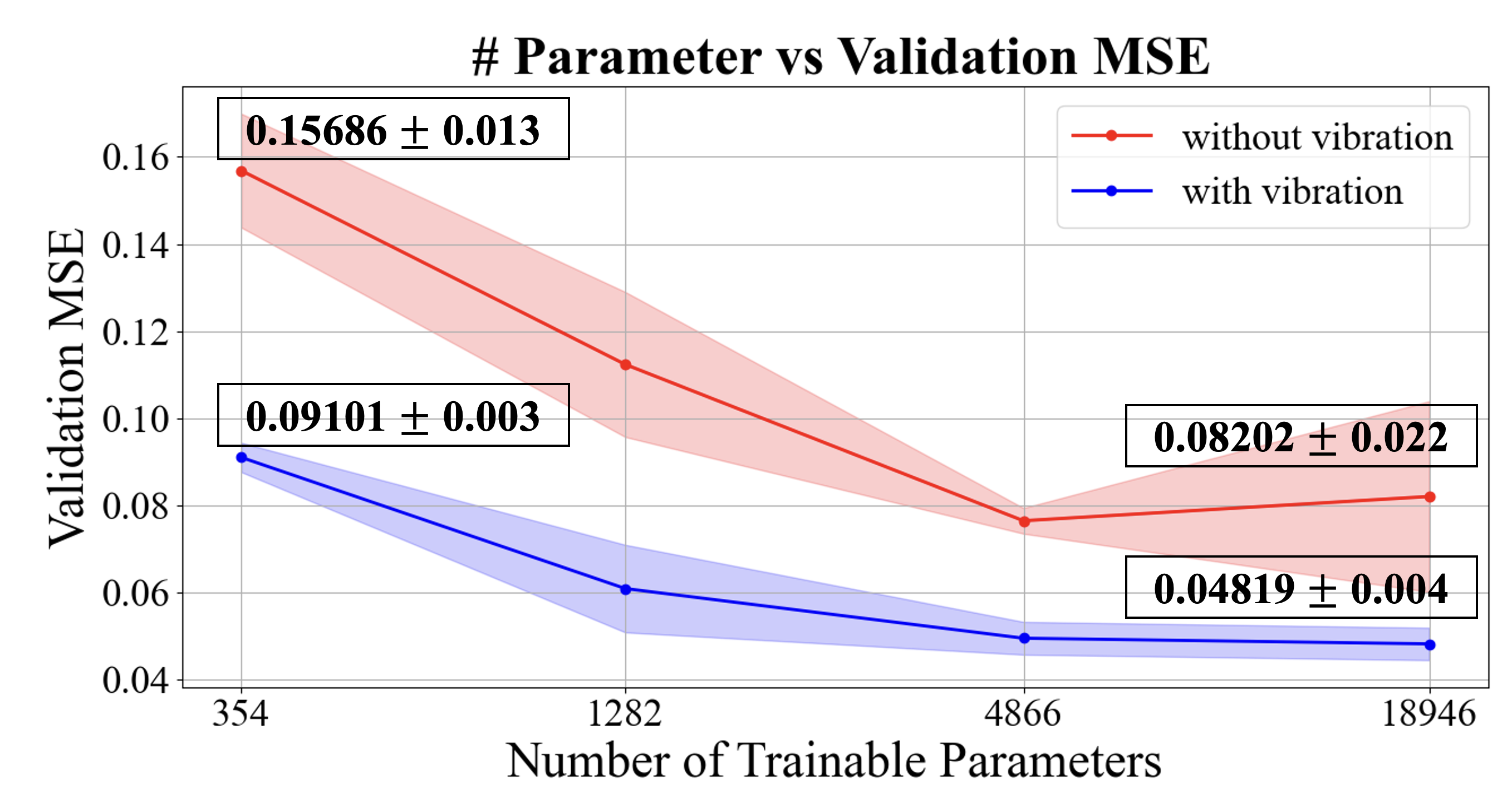} 
    \caption{MSE loss of the TCN model on the unseen validation set, representing the estimation performance for predicting the command position ($\textbf{p}_\text{cmd}$) based on the measured position history ($\textbf{p}_\text{meas}$). The red line corresponds to the TCN model trained without vibration, while the blue line represents the model trained with vibration. The shaded region indicates the standard deviation of MSE across three training runs.}
     
    \label{fig:model_param}
\end{figure}

\begin{figure*}[t!]
    \centering
    \vspace{2mm}
    \includegraphics[width=0.9\textwidth]{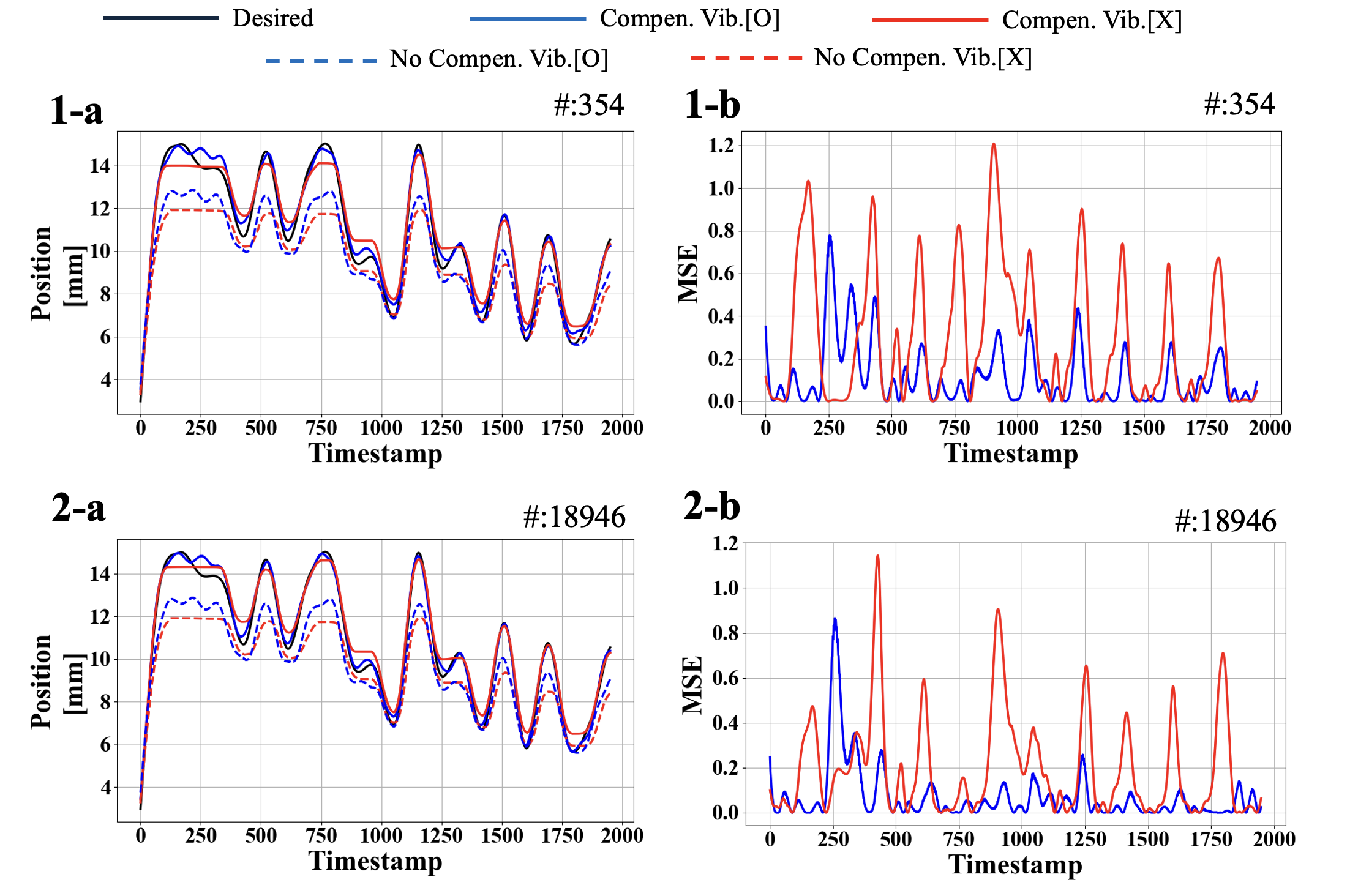} 
    \caption{Comparison of compensation results for different TCN model sizes. (1-a) Position tracking over time and (1-b) mean squared error (MSE) analysis with and without vibration using the 354-parameter TCN model. (2-a) and (2-b) present the same analyses for the 18,946-parameter TCN model.} 
    \label{fig:comp_result} 
\end{figure*}

\subsection{Deep Learning-Based Modeling}
\label{sec:deep_learning}
We employed a Temporal Convolutional Network (TCN) \cite{tcn} for hysteresis estimation due to its proven effectiveness in tendon-driven manipulators \cite{park2024hysteresis}. TCN efficiently compensates for the hysteresis effect by leveraging historical information, as hysteresis exhibits time-dependent properties. The TCN architecture consists of serially connected residual blocks, each comprising two dilated convolutions, two weight normalization layers, and two ReLU activation functions. The dilated convolutions are performed as follows:
\begin{gather}
F_n(s) = (x *_d f)(s) = \sum_{i=0}^{k-1} f(i) \cdot x_{s - d \cdot i}
\end{gather}

where \( k \) denotes the kernel size, \( n \) is the index of the residual block, \( f \) represents the convolution filter, and the number of filters in each convolution layer is equivalent to \( C_{\text{out}} \). The sequence length is denoted by \( s \), and \( d \) represents the dilation factor. The residual blocks utilize exponentially increasing dilation factors with a base of 2 (i.e., \( 2^0, 2^1, 2^2, ..., 2^{n-1} \)). The total number of residual blocks (\( n \)) is determined as follows:
\begin{gather}
nb = \text{num\_block} = \left\lceil \log_{2} \frac{(L-1)}{2k-2} + 1 \right\rceil
\end{gather}

The number of trainable parameters in the TCN can be approximated as:
\begin{gather}
P = 2 \sum_{n=1}^{nb-1} \left( c_{\text{out}, \:n-1} \cdot c_{\text{out},\:n} \cdot k \right) + \left( c_{\text{in}} \cdot c_{\text{out}, \:0} \right)
\end{gather}

where \( c_{\text{out}, n} \) represents the number of filters in the \( n \)th residual block. Using this TCN, we model the hysteresis effect in the tendon-sheath mechanism (TSM). The input to the model consists of the historical measured position values, while the output corresponds to the estimated commanded joint angles:
\begin{gather}
\hat{\textbf{p}}_\mathrm{cmd}^{(t)} =  f_{\theta} (\textbf{p}_\mathrm{meas}^{(t-L, \: t-L+1, \:..., \:t-1, \:t)})
\label{eq:tcn}
\end{gather}

where \( L \) is the length of the input sequence. In our implementation, we set \( L = 40 \), meaning that the model estimates the current commanded position using 720 ms of historical data (\( 40 \times 18 \) ms).

For training, we utilized the datasets \( \mathcal{D}_{\text{no vib}} \) and \( \mathcal{D}_{\text{vib}} \). A total of 12,000 samples were allocated for training, while 3,000 samples were used for validation. By deploying the trained TCN model on both the non-vibration dataset (\( f^{\text{no vib}}_{\theta} \)) and the vibration-applied dataset (\( f^{\text{vib}}_{\theta} \)), we implemented a compensation control loop:
\begin{gather}
\hat{\textbf{p}}_\mathrm{cal \: cmd, vib}^{(t)} =  f^{\text{vib}}_{\theta} (\textbf{p}_\mathrm{desired}^{(t-L, \: t-L+1, \:..., \:t-1, \:t)}) \\
\hat{\textbf{p}}_\mathrm{cal \: cmd, no \: vib}^{(t)} = f^{\text{no vib}}_{\theta}  (\textbf{p}_\mathrm{desired}^{(t-L, \: t-L+1, \:..., \:t-1, \:t)})
\label{eq:tcn_inference}
\end{gather}

where \( \textbf{p}_\text{desired} \) denotes the desired position, while \( \hat{\textbf{p}}_\mathrm{cal \: cmd, vib} \) and \( \hat{\textbf{p}}_\mathrm{cal \: cmd, no \: vib} \) represent the calibrated command position values that allow the system to reach the desired position accurately.

\subsection{Ablation Study on the Number of Trainable Parameters}
To analyze the effect of model complexity on performance, we conducted an ablation study by varying the number of trainable parameters in the deep learning model. The study evaluates the model’s performance under both vibration-applied and non-vibration conditions. Adjusting the number of trainable parameters affects the model's capacity to capture the hysteresis effect. 

We trained the TCN model with four different parameter configurations: 354, 1,282, 4,866, and 18,946. Training was conducted separately on both the vibration-applied dataset (\(\mathcal{D}_{\text{vib}}\)) and the non-vibration dataset (\(\mathcal{D}_{\text{no vib}}\)), maintaining consistent training conditions across all settings. This resulted in a total of eight models (four different parameter settings for both vibration and non-vibration cases).

Each trained model was evaluated on the validation dataset, and the results are summarized in \Cref{fig:model_param}. Despite being trained under identical conditions except for dataset selection, models trained on the vibration-applied dataset demonstrated significantly improved performance. For the smallest model (354 parameters), the mean squared error (MSE) for $f^\text{no vib}_{\theta}$ was 0.1596, whereas for $f^\text{vib}_{\theta}$, it was 0.091, reflecting an approximately 43.0\% reduction in error. Similarly, for the largest model (18,946 parameters), the MSE values were 0.082 and 0.048 for $f^\text{no vib}_{\theta}$ and $f^\text{vib}_{\theta}$, respectively, showing a 41.5\% reduction. 

This result indicates that the dataset with vibration exhibits lower complexity and is easier to model. Additionally, $f^\text{no vib}_{\theta}$ shows a higher standard deviation (as represented by the width of the line) across three training runs, suggesting that it possesses highly nonlinear properties and is more sensitive to initial weight initialization. Notably, the vibrating data trained with the smallest model (354 parameters) achieves comparable performance to the non-vibrating data trained with the largest model (18,946 parameters), implying that vibration significantly reduces the nonlinearities in the TSM.

\begin{table}[t!]
\centering
\caption{Comparison of Mean Absolute Error (MAE) and standard deviation (STD) between $\textbf{p}_{cmd}$ and $\textbf{p}_{meas}$ across different model sizes, with and without vibration, for compensated and non-compensated control. \# states the number of trainable parameters}
\label{tab:comp_result}
\resizebox{0.8\linewidth}{!}{%
\begin{tabular}{cccc} 
\toprule
\textbf{Method}                                                               & \textbf{Vibration} & \begin{tabular}[c]{@{}c@{}}\textbf{MAE}\\{[}mm]\end{tabular} & \begin{tabular}[c]{@{}c@{}}\textbf{STD}\\{[}mm]\end{tabular}  \\ 
\hline
\multirow{2}{*}{\begin{tabular}[c]{@{}c@{}}No\\Compen\end{tabular}}           & \ding{55}                 & \textbf{\textbf{1.334}}                                      & 0.964                                                         \\
                                                                              & \ding{51}                 & \textbf{\textbf{1.077}}                                      & 0.692                                                         \\ 
\hline
\multirow{2}{*}{\begin{tabular}[c]{@{}c@{}}Compen.\\(\#: 354)\end{tabular}}   & \ding{55}                 & \textbf{0.4336}                                              & 0.300                                                         \\
                                                                              & \ding{51}                 & \textbf{0.2757}                                              & 0.184                                                         \\ 
\hline
\multirow{2}{*}{\begin{tabular}[c]{@{}c@{}}Compen.\\(\#: 1282)\end{tabular}}  & \ding{55}                 & 0.3719                                                       & 0.253                                                         \\
                                                                              & \ding{51}                 & 0.1984                                                       & 0.162                                                         \\ 
\hline
\multirow{2}{*}{\begin{tabular}[c]{@{}c@{}}Compen.\\(\#: 4866)\end{tabular}}  & \ding{55}                 & 0.3913                                                       & 0.286                                                         \\
                                                                              & \ding{51}                 & 0.2200                                                       & 0.195                                                         \\ 
\hline
\multirow{2}{*}{\begin{tabular}[c]{@{}c@{}}Compen.\\(\#: 18946)\end{tabular}} & \ding{55}                 & \textbf{0.3703}                                              & 0.257                                                         \\
                                                                              & \ding{51}                 & \textbf{0.1969}                                              & 0.157                                                         \\
\toprule
\end{tabular}
}
\end{table}

\section{Results}
This section presents the experimental evaluation of the proposed hysteresis compensation approach. The primary focus is to assess the effectiveness of the trained model in mitigating hysteresis under two conditions: with and without applied vibrational excitation.

\subsection{Hysteresis Compensation Performance}
To evaluate compensation performance, we applied the trained models from \eqref{eq:tcn} to an unseen random desired trajectory of length 3500, generated using the approach described in  \Cref{sec:random_trajectory}. For the vibration-trained model, a 70 Hz excitation frequency was maintained during inference to match training conditions.

The results are summarized in \Cref{fig:comp_result} and \Cref{tab:comp_result}. Without compensation, the mean absolute error (MAE) in the no-vibration condition was 1.334 mm. In contrast, compensation using the vibration-trained model reduced the MAE to 0.1969 mm, corresponding to an 85.2\% reduction in hysteresis. 

\Cref{tab:comp_result} further compares compensation performance across different model complexities, highlighting the impact of parameter size. Notably, the smallest TCN model (354 parameters) trained with vibration (0.2757 mm MAE) outperformed the largest TCN model (18,946 parameters) trained without vibration (0.3703 mm MAE). These results suggest that the vibrational dataset exhibits lower nonlinearities, making the system dynamics more predictable and easier to model. Additionally, vibration mitigates static friction and dead-zone effects, facilitating more effective compensation. These findings underscore the potential applicability of vibration-enhanced compensation strategies in tendon-driven continuum manipulators.

\section{CONCLUSIONS}
This study demonstrates that controlled vibrational motion significantly reduces hysteresis in tendon-sheath mechanisms (TSMs), improving trajectory tracking accuracy and enabling more efficient compensation strategies. Experimental results confirm that applying vibration decreases RMSE between the desired and measured positions by up to 23.41\% (2.2345 mm → 1.7113 mm) on random trajectories while enhancing correlation, indicating that reducing dead zones improves system responsiveness.

To further validate the effects of vibration, we analyzed TCN-based modeling with varying trainable parameters (354, 1,282, 4,866, and 18,946). Results show that a vibration-assisted dataset with only 354 parameters achieves similar estimation performance to a non-vibration dataset with 18,946 parameters, demonstrating that without vibration, the system is harder to model and requires greater complexity.

When integrated with a TCN-based compensation model, vibration improves performance, achieving an 85.2\% reduction in MAE (1.334 mm → 0.1969 mm). In comparison, without vibration, the TCN-based approach reduces MAE by 72.1\% (1.334 mm → 0.3703 mm). These results highlight that vibration reduces system nonlinearities and, when combined with data-driven modeling, effectively mitigates hysteresis effects.

By introducing vibration, hysteresis in TSMs is fundamentally reduced, simplifying compensation modeling and lowering RMSE by mitigating static friction between the tendon and sheath. These findings provide a promising and practical solution for TSM-based robotic applications. Future research will focus on understanding the underlying mechanisms by which vibration reduces nonlinearities and RMSE. Additionally, we will explore the application of vibration-assisted mechanisms in flexible robots and assess its broader effects on system performance.

\section{acknowledgement}
This work was supported by Korea Medical Device Development Fund grant funded by the Korea government (1711196477, RS‐2023‐00252244), the National Research Council of Science \& Technology (NST) grant funded by the Korea government (MSIT) (CRC23021‐000), the Industrial Strategic Technology Development Program (ISTDP) (RS-2024-00443054) funded by the Ministry of Trade, Industry \& Energy (MOTIE, Korea), and by the collaborative project with ROEN Surgical Inc.



\bibliographystyle{IEEEtran}

\bibliography{ref}
\end{document}